\documentclass{article}

\usepackage[preprint,nonatbib]{nips_2018}

\usepackage{graphicx}
\usepackage{multirow}
\usepackage{amsmath,amssymb} 
\usepackage[utf8]{inputenc} 
\usepackage[T1]{fontenc}    
\usepackage{hyperref}       
\usepackage{url}            
\usepackage{booktabs}       
\usepackage{amsfonts}       
\usepackage{nicefrac}       
\usepackage{microtype}      

\title{Self-supervisory Signals for Object Discovery and Detection}

\author{
  Etienne Pot\thanks{Work done as a member of the Google Brain Residency (g.co/airesidency)} \\
  Google Brain \\
  \texttt{epot@google.com} \\
  \And
  Alexander Toshev \\
  Google Brain \\
  \texttt{toshev@google.com} \\
  \And
  Jana Kosecka \\
  Google Brain \\
  \texttt{kosecka@google.com} \\
}

\begin{document}

\maketitle

\begin{abstract}
In robotic applications, we often face the challenge of discovering new objects while having very little or no labelled training data. 
In this paper we explore the use of self-supervision provided by a robot traversing an environment to learn representations of encountered objects. Knowledge of ego-motion and depth perception enables the agent to effectively associate multiple object proposals, which serve as training data for learning object representations from unlabelled images. 
We demonstrate the utility of this representation in two ways. First, we can automatically discover objects by performing clustering in the learned embedding space. Each resulting cluster contains examples of one instance seen from various viewpoints and scales. Second, given a small number of labeled images, we can efficiently learn detectors for these labels. In the few-shot regime, these detectors have a substantially higher mAP of $0.22$ compared to $0.12$ of off-the-shelf standard detectors trained on this limited data. Thus, the proposed self-supervision results in effective environment specific object discovery and detection at no or very small human labeling cost.
\end{abstract}

\section{Introduction}

Traditionally, object detection systems are trained on large datasets, labeled with bounding boxes from a pre-defined set of common categories. This has been proven highly successful for applications where the object types are known and ample labeled data is available for them.  In robotic applications, however, often a robot needs to detect novel objects with no or limited supervision or adapt current models to new environments.  
In this work, we propose to leverage the motion of a robotic agent exploring an environment as main means of self-supervision, and show how to discover novel objects and learn effective detection models for them with no or very limited labelling effort.

\begin{figure}
    \centering
    \includegraphics[width=0.45\textwidth]{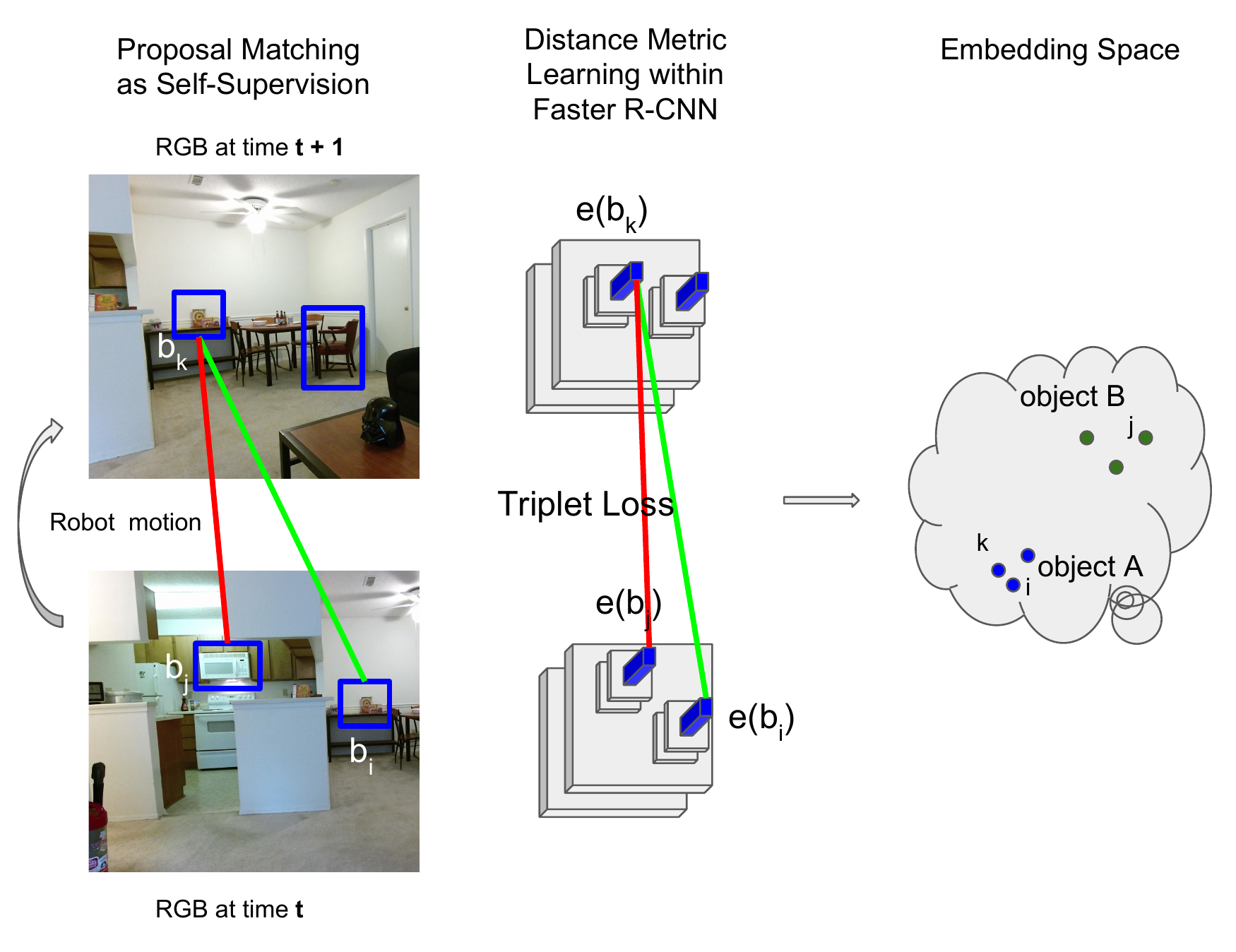}\hfill
    \includegraphics[width=0.5\textwidth]{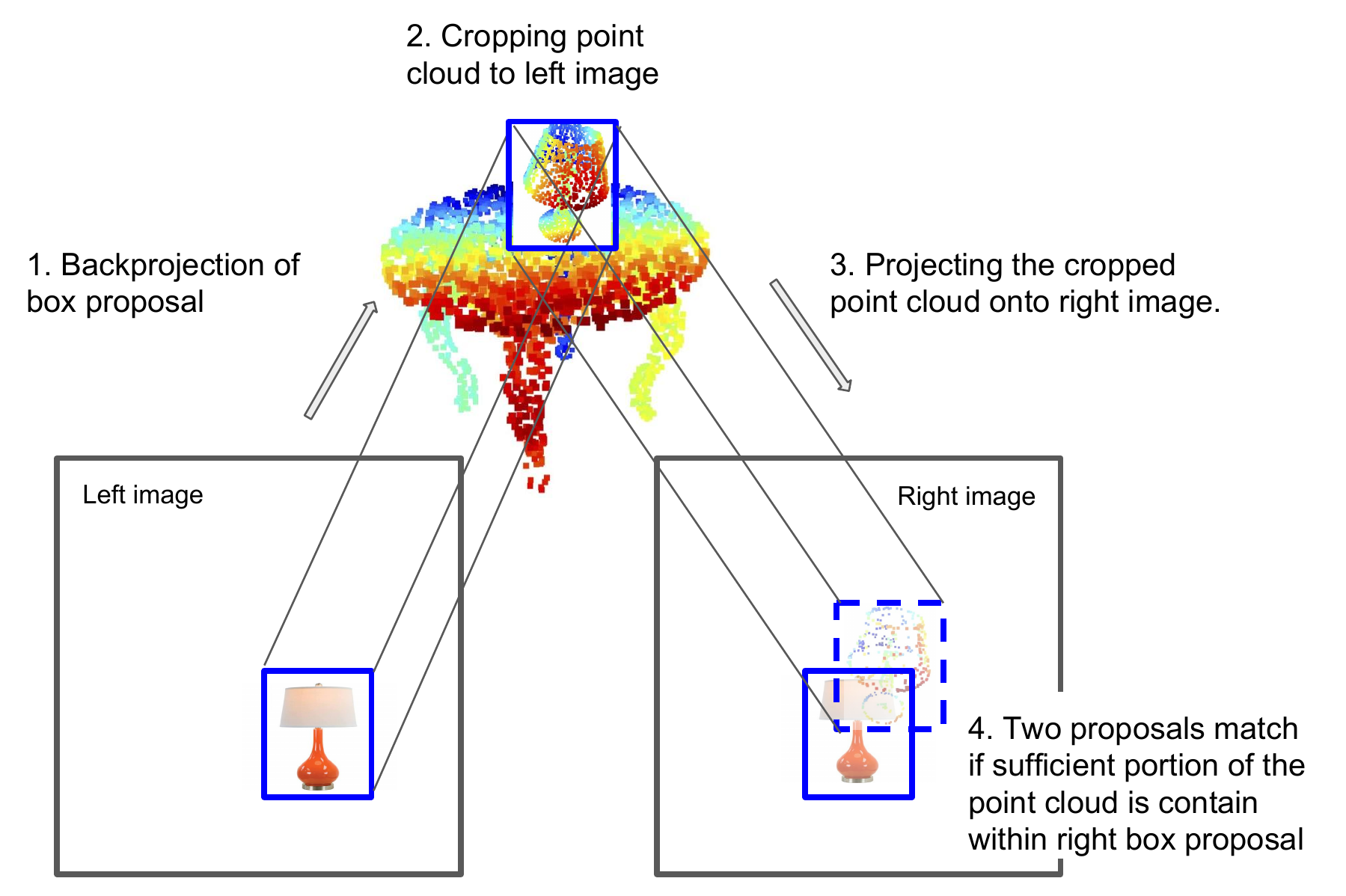}
    \caption{Left: Overview of the approach. Right: Self-supervised data association.}
    \label{fig:intro}
\end{figure}

We start by learning an \textit{embedding of objects} encountered in the environment during exploration by establishing correspondences  between object proposals produced by a generic class agnostic detector (see Fig.~\ref{fig:intro}, left). This can be done without any human supervision using ego-motion and depth cues. These associations are used for training an object representation via a Distance Metric Learning loss~\cite{schroff2015facenet} integrated as part of Faster-RCNN detector~\cite{ren2015faster}. 
The learned representation is subsequently utilized for \textit{unsupervised object discovery} and \textit{few-shot object detection} (see Fig.~\ref{fig:intro}, right). 
Using clustering techniques in the learned embedding space, we obtain groupings containing examples of encountered object instances. Those clusters can then be labeled to train an object detector.
In this manner our method discovers new objects and provides training examples for object detection at no or very small labelling cost. 

We propose and evaluate two object detection approaches under the umbrella name of Self-Supervision based Object Detection (SSOD). For the first approach, called SSOD-Cluster, we train a Faster-RCNN model on the discovered clusters. This method requires mere assigning labels to the produced clusters without having to draw bounding boxes. The second approach, called SSOD-Dist, leverages nearest-neighbor based classification. For each object proposal at test time, we find the closest embedding in a small set of bounding boxes labelled in training images.

Both approaches result in highly superior detectors compared to off-the-shelf Faster RCNN model fine-tuned on similarly sized labeled data. In particular, SSOD-Cluster results in mean Average Precision (mAP) of $0.46$ while the Faster RCNN fine-tuned on similar number of object examples has mAP of $0.38$. While standard detector requires bounding boxes, SSOD-Cluster only requires labels for clusters as a human supervision. Further, SSOD-Dist, when given a single training example had an mAP of $0.23$ vs $0.12$ for Faster RCNN. This demonstrates the utility of self-supervision for object detection. The proposed approach is an effective way to train performant detection models for newly discovered models at no or very small labelling effort.  

\section{Related Work}

\paragraph{Unsupervised / Weakly Supervised Object Discovery} Automatic object discovery in image collections has long history in computer vision \cite{sivic2005discovering,russell2006using}. Earlier approaches are usually based on object proposals, obtained via bottom-up perceptual grouping mechanisms followed by latent topic discovery. 
The training and evaluation of these methods was applied to image collections, where there are no relations among the images. 

In environments relevant to robotic applications, earlier approaches for object discovery used strategies for generating category independent object proposals~\cite{Uijlings_IJCV13} or clustering local features followed classification~\cite{Collet_ICRA16}. In the RGB-D settings~\cite{Mishra_ICRA12} used object boundaries to guide the detection of fixation points denoting the presence of objects, while~\cite{Karpathy_ICRA13} performed object discovery
by ranking 3D mesh segments based on objectness scores. These methods typically used motion, depth and contextual cues, but the detection and categorization was followed using hand-engineered features and worked well mostly on textured objects. 

The current state of the art object detectors~\cite{ren2015faster,ssd_eccv2016} that build on successes of deep convolutional neural network (CNN) architectures require large amounts of training data and typically focus on commonly encountered object categories in independent images. To offset the amount of needed supervision, several  
approaches rely on videos, where object hypotheses can be tracked \cite{kwak2015unsupervised,pathak2017learning,schulter2013unsupervised,agrawal2015learning,wang2015unsupervised,wang2014video}. 
Contrary to our setup, these methods expect that the video contains one dominant object present in most of it.
Pathak et al.~\cite{pathak2017learning} also use videos to learn 
more general representations with supervision coming from motion segmentation. 
Authors in Agrawal et al.~\cite{agrawal2015learning}, trained representation to predict the ego-motion and demonstrated that the learned features compare favorably to features learned by class-label supervision on some specific tasks. However, the above approaches do not learn environment specific representations,
and consider the full image instead of individual objects.

Similar to our model is the triplet formulation by Wang and Gupta~\cite{wang2015unsupervised} where the training examples are generated by tracking patches in unlabelled videos. Yet this feature representation is on full images and is not specific to one environment. 

\paragraph{Self-Supervised Learning} Another relevant stream of work uses self-supervision \cite{pathak2017curiosity,pinto2016supersizing,mitash2017self,hickson2018object,sermanet2017time} to offset the need for labelled training data.  Self-supervision in robotic setting pertains to capability of learning effective representations of sensory streams by generating actions in a goal agnostic manner and optimizing goal agnostic learning objectives.  Successful applications of this idea have been found in robotics for grasping and manipulation \cite{sermanet2017time,pinto2016supersizing} and navigation \cite{pathak2017curiosity}. Pathak et al.~\cite{pathak2017curiosity} proposes to predict the result of robot actions along with the prediction of the effect of that action.  More similarly to us, Sermanet at el.~\cite{sermanet2017time} uses a triplet loss to utilize same / not same events in video to learn view independent embeddings of human actions. The similarity in their work represents events observed from different views on a full image scale, while in our approach we focus on objects.

The closest to our work is the applications of self-supervision to detection. Mitash et al.~\cite{mitash2017self} trains detectors in simulation and improves them on real unlabeled data, where scenes are observed from different viewpoints. By relating detections across different viewpoints, the authors generate labels for the real images, which are used to fine-tune the detector. Their model, however, does not focus on developing a general representation for discovery and detection. Finally, Hickson et al.~\cite{hickson2018object} develops a clustering training procedure as part of a ConvNet training, which can be applied, among other things, to discover novel objects in videos.
 
Another line of related work deals with few-shot object detection and categorization. 
In robotic settings, recent work by~\cite{Held_ICRA16} demonstrated that effective object representation invariant to pose and background variations can be learned from multi-view object datasets with classification loss and can be then quickly adapted to new instances using only a single training example. This approach has been tested on object instance categorization tasks and has not been extended to object detection, where objects often exhibit large scale variations. 
\section{Self-Supervision for Object Detection}
The detector we use is built upon Faster-RCNN~\cite{ren2015faster} trained on the COCO object detection dataset~\cite{lin2014microsoft}. This model is a two-branch network, where the first branch predicts a number of class agnostic object proposals which are further classified by the second branch. Both branches share the same feature extraction sub-network.
In this work, the representation learning is done by modifying the second branch so that it outputs a vectorial embedding instead of a probability distribution over classes. Thus, the object representation learning is integrated as part of a state-of-art detection algorithm, and inherits its computational efficiency.

\subsection{Self-Supervisory Signal}  \label{sec:self_supervisory_signal}
The self-supervisory signal is based on the ability to relate object proposals across frames captured by the robot during exploration of an environment as show in Figure~\ref{fig:intro}. More precisely, denote by $b_i^k$ the $i^\textrm{th}$ object proposal from the $k^\textrm{th}$ frame. Then we define a self-supervision similarity $ss$ of two proposals from two different frames $k$ and $l$ as:
\begin{equation}\label{eq:ss_overview}
ss(b_i^k, b_j^l) = 1\textrm{ iff proposals }i\textrm{ and }j\textrm{ denote the same object, otherwise }0
\end{equation}

In order to compute the association of bounding boxes we assume 
that agent can estimate locally its pose. This can be achieved through Simultaneous Mapping and Localization SLAM~\cite{durrant2006simultaneous}, a capability which is ubiquitous across robotic systems. The pose of the $k^\textrm{th}$ camera is characterized by rotation $R_k$ and translation $t_k$ w.r.t. a fixed coordinate system. We also assume that depth of the scene for each frame in the form of a point cloud is available. The depth can be obtained either by depth sensors (such as LiDAR) or estimated using stereo. Denote $P_k$ the point cloud associated with $k^\textrm{th}$ frame.

The estimates of motion and depth enable us to associate bounding boxes generated by object proposal part of Faster-RCNN network. 
Two object proposals generated in $k^\textrm{th}$ and $l^\textrm{th}$ frames represent the same object instance if the reprojection of $b_i^k$
via 3D to frame $l$ overlaps significantly with $b_j^l$
(see Fig.~\ref{fig:intro}, right). More precisely, denote by $p_i^k$ the subset of the point cloud associated with image $k$ which when projected is completely contained in proposal bounding box $b_i^k$. We can transform this point cloud into the coordinate system of a second image $l$ and project it using the following operator:
\[
    T_{k\rightarrow l}(p_i^k) = \{K(T_l^{-1}(T_k(x)))\textrm{ for }x\in p_i^k\}
\]
where $T_l=[R_l|t_l]$ is the camera pose of $l^\textrm{th}$ image as defined above and $K$ is the intrinsic camera parameters matrix. We then estimate the coordinates of $b_i^{k \rightarrow l}$, the bounding box $b_i^k$ projected on image $l$, by taking the minimum and maximum $x$ and $y$ coordinates of $T_{k\rightarrow l}(p_i^k)$. Then the two object proposals match iff the intersection over union ($IoU$) between the projected bounding box $b_i^{k \rightarrow l}$ and $b_j^l$ is above a threshold $th$. In our experiments we set ${th}=0.1$. If more than two bounding boxes match together, only the pairs with the highest $IoU$ score are kept.

\subsection{Representation Learning}
To learn a vectorial embedding of objects, we use the above self-supervision to generate training examples for a distance metric learning approach. We employ a triplet loss~\cite{schroff2015facenet}, although any distance metric learning loss is applicable.
To define a training objective one needs to define a set of triplets, each representing similar and dissimilar object proposals. Similar proposals are obtained by self-supervision described in Section \ref{sec:self_supervisory_signal}. Dissimilar ones can be picked from the same frame -- if they are not overlapping, then they are guaranteed to represent different objects.
For an object proposal $b_i^k$, called reference, one can choose another proposal $b_j^l$ from a different image such that $ss(b_i^k, b_j^l) = 1$. A third proposal $b_{j'}^k$ is chosen from the same frame $k$ as the reference. Then, for a set of such triplets the loss reads:
\begin{align}
    Loss = \sum_{ss(b_i^k, b_j^l) = 1, j\neq j'}\max\{||e(b_i^k) - e(b_j^l)|| - ||e(b_i^k) - e(b_{j'}^k)|| + M, 0\}
\end{align}
where $e(b)$ is the embedding vector generated by the detector for bounding box proposal $b$ and $M$ is a margin value. In our implementation we set $M=1.0$.

Note that the above embedding is trained using frames captured from one particular environment. As it does not require human supervision, the robot can generate an ample amount of data by roaming around. In our implementation, we freeze the feature extraction sub-network and the region proposal network of the detector and only fine-tune the classifier of the second stage. The last SoftMax layer is replaced by a fully-connected projection followed by a L2-normalization.
The resulting embedding is specific to the objects in the environment where the training data was generated, which can be thought of as an adaptation procedure for that particular environment. 

\textbf{Training} The network is trained using Adam optimizer for 35,000 steps. At each step, a location is randomly sampled and its neighbouring locations are used for bounding box matching. Thus each step correspond to a batch of 24 images/locations. Each image contains a different number of bounding box proposals so the number of embeddings per batch varies between 100 and 300. The network is trained using a learning rate of 0.0001 with decay of 0.94. The embedding dimension is chosen to be 128. 

\subsection{Applications} \label{object_discovery}
\textbf{Object Discovery} The learned embedding space for one particular environment can be used to discover new object instances. Since the learning objective is for the embeddings of same instances to lie nearby, as shown in Fig.~\ref{fig:tsne}, clustering can be used to 
group the embeddings. We use Mean Shift clustering \cite{comaniciu2002mean} on the embeddings of object proposals from one environment. This algorithm produces clusters or modes, which represent  examples of encountered objects in multiple viewpoints.  An example of such clusters can be seen in Fig.~\ref{fig:clusters}. In the experimental section we quantify the result of this clustering. 

\label{few_shot}
\textbf{Object Detection} Another application of the embedding space is object detection, which we call Self-supervision based Object Detection (SSOD). We use two variants. SSOD-Cluster is based on Faster-RCNN detector, which is trained by assigning labels to the entire clusters and using the proposal bounding boxes belonging to the cluster as training examples. 

In the second variant, called SSOD-Dist we label a small set of bounding boxes
$D=\{(b_i, y_i)\}_{i=1}^n$, where $y_i$ represents a class label, and classify the object proposals using nearest-neighbor classification procedure, similar to \cite{schroff2015facenet}.
We classify a query proposal $b$ as follows:
\[
  y_{i^*}\textrm{ for }i^*=\arg\min_i\{||e(b_i) - e(b)||\}
\]
The embedding $e(\cdot)$ produced by the detector is normalized. We show in the experimental section, when the size of the training data $D$ is small, the above procedure results in higher detection results, measured by mean average precision. 
\section{Evaluation}  \label{sec:evaluation}

\paragraph{Dataset} We use the Active Vision Dataset (AVD) \cite{ammirato2017dataset} for training and evaluation of the proposed approach. The dataset consists of dense scans of  $9$ different home environments. 
Each house has observations captured at grid of locations with RGB-D images at $6$ different orientations and all the images registered with respect to a global coordinate frame. The dataset enables virtual traversals of each environment with camera poses and RGB-D views associated with every observation. The scenes contain $33$ object instances of common household products including bottles, cans, cereals boxes, etc. The object instances are a subset of objects from BigBird dataset~\cite{bigbird2014} and most of them do not belong to the classes of the pre-trained detector~\cite{ren2015faster}.
Five of those environments were captured twice with different placement of the objects. Thus, for each those environments one can use the first scan for discovery and detector learning, and the second for evaluation. Because the objects are small graspable products, we remove for training and testing far-away objects with bounding boxes smaller than $100\times 75$, following \cite{ammirato2017dataset}.

\begin{figure}
    \centering
    \includegraphics[width=0.45\textwidth]{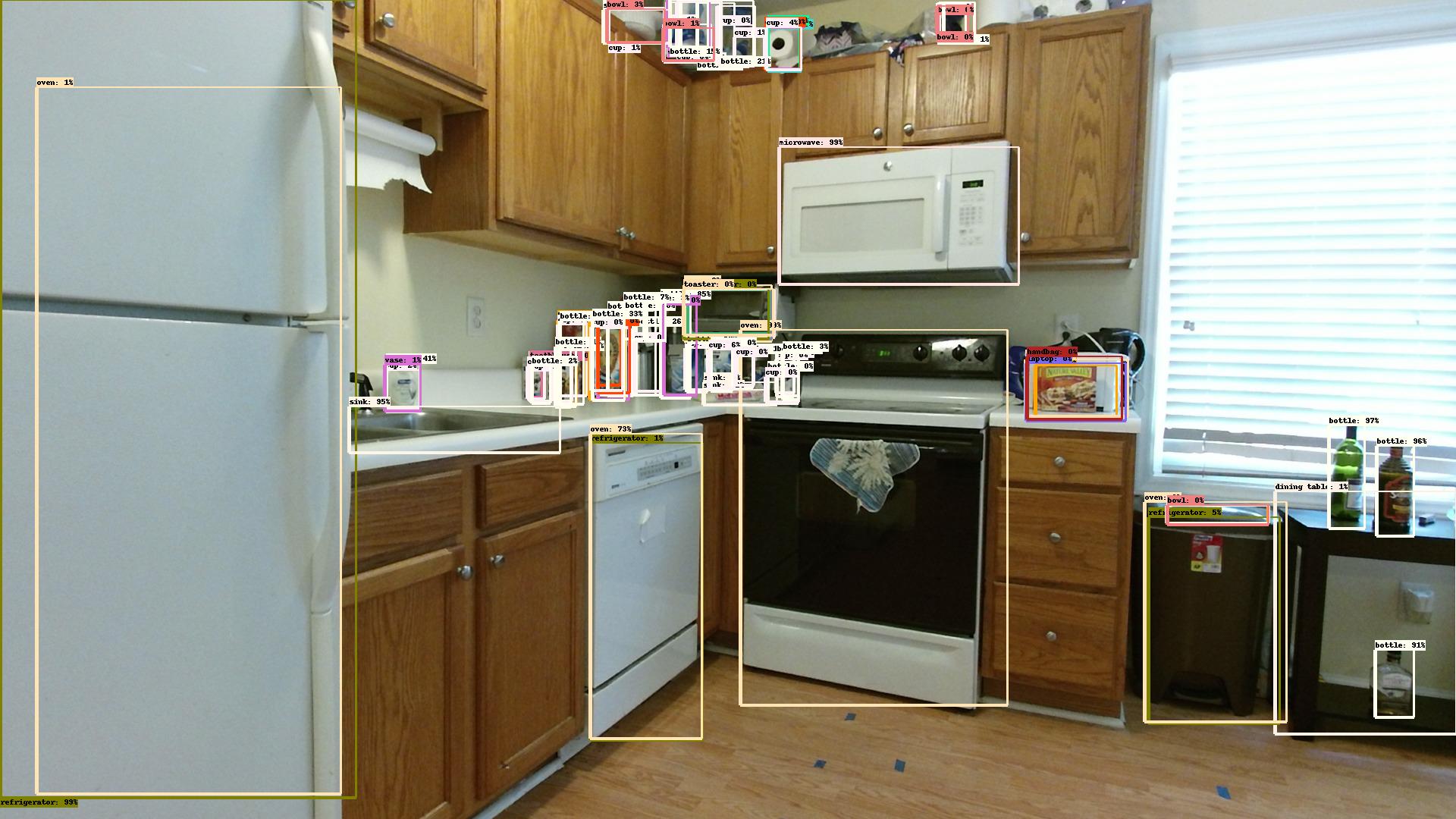}
    \includegraphics[width=0.45\textwidth]{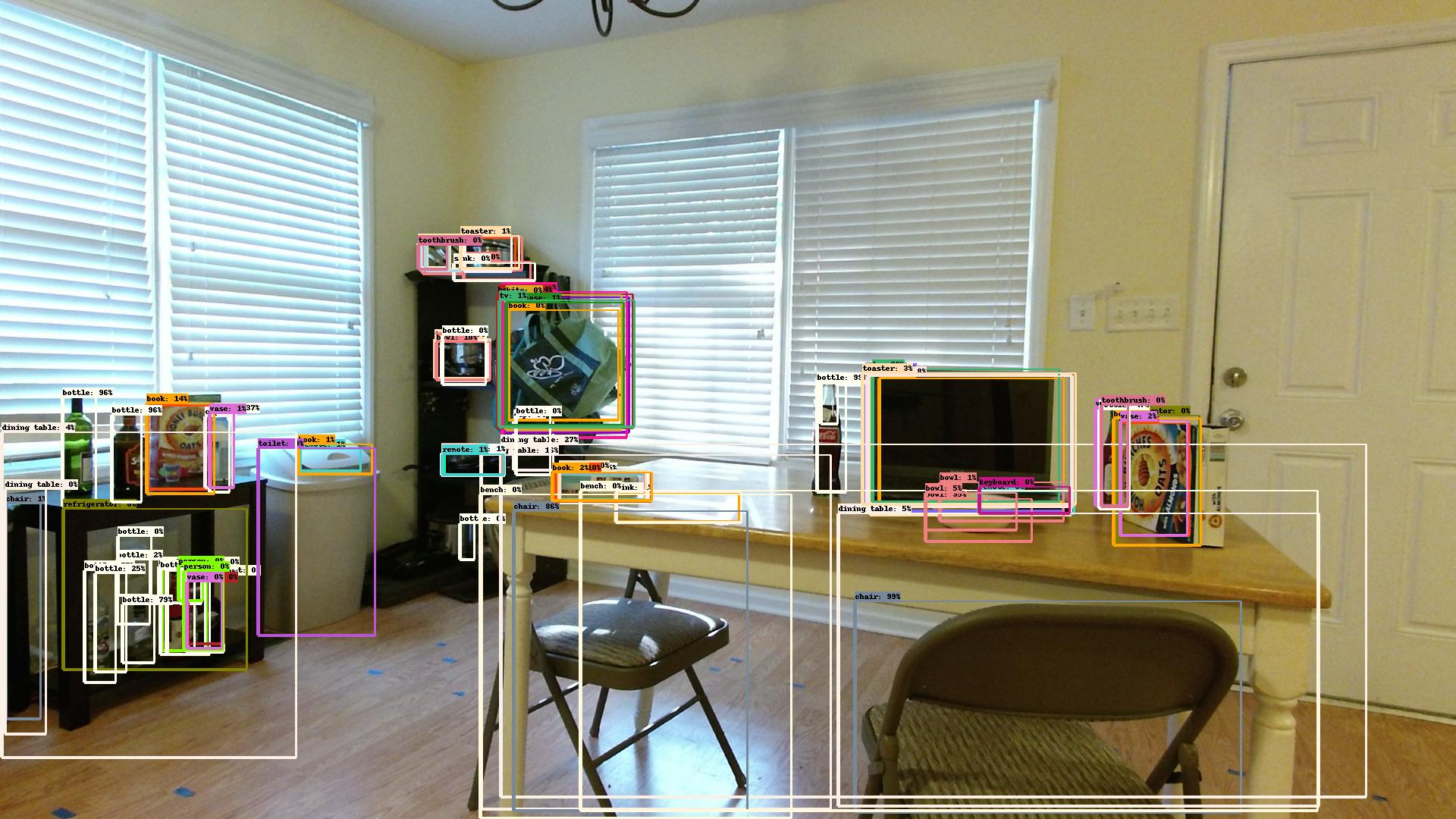}
    \caption{Example output of the COCO-trained proposal generator on Active Vision Dataset~\cite{ammirato2017dataset} images.}
    \label{fig:coco_proposals}
\end{figure}

\paragraph{Object Proposals} We use Faster-RCNN pre-trained with ResNet-101 on COCO dataset. In order to have the proposals cover a large set of objects (including the classes not present in the original COCO dataset), we reduce the objectness threshold score of Region Proposal Network (RPN) to $0.01$ and apply non-maximum suppression with an IoU threshold of $0.7$ to remove bounding box duplicates (see Fig.~\ref{fig:coco_proposals}).
To verify that the proposal generator captures AVD objects, we measure the recall of the COCO-trained RPN on AVD, where the recall has been measured per objects category:

\begin{center}
\begin{tabular}{|c|c|c|c|c|}
    \hline
    cereal box & can & soap & bottle & other \\
    \hline
    0.51 & 0.55 & 0.70 & 0.68 & 0.70 \\
    \hline
\end{tabular}
\end{center}
Note that for instances such as bottles, the RPN has a recall around $\geq 0.68$, while the cereal boxes recall is much lower. Hence, some of the instances are well captured by the generated proposals, while others will not be discoverable by our approach. As shown in Table \ref{tab:map_per_instance}, the recall scores correlate with final performance.

\subsection{Object Discovery}  \label{sec:obj_disc}

For the purpose of object discovery, we train an object representation and use it to cluster the object proposals, as outlined in Sec.~\ref{object_discovery}. 
Clustering enables us to discover object  instances without any human labelling. We use Mean-Shift clustering with different bandwidths to avoid specifying the number of clusters and filter clusters with less than 8 images. For example, a bandwidth of 0.6 produces 176 clusters.

\begin{figure}
    \centering
    \includegraphics[width=0.9\textwidth]{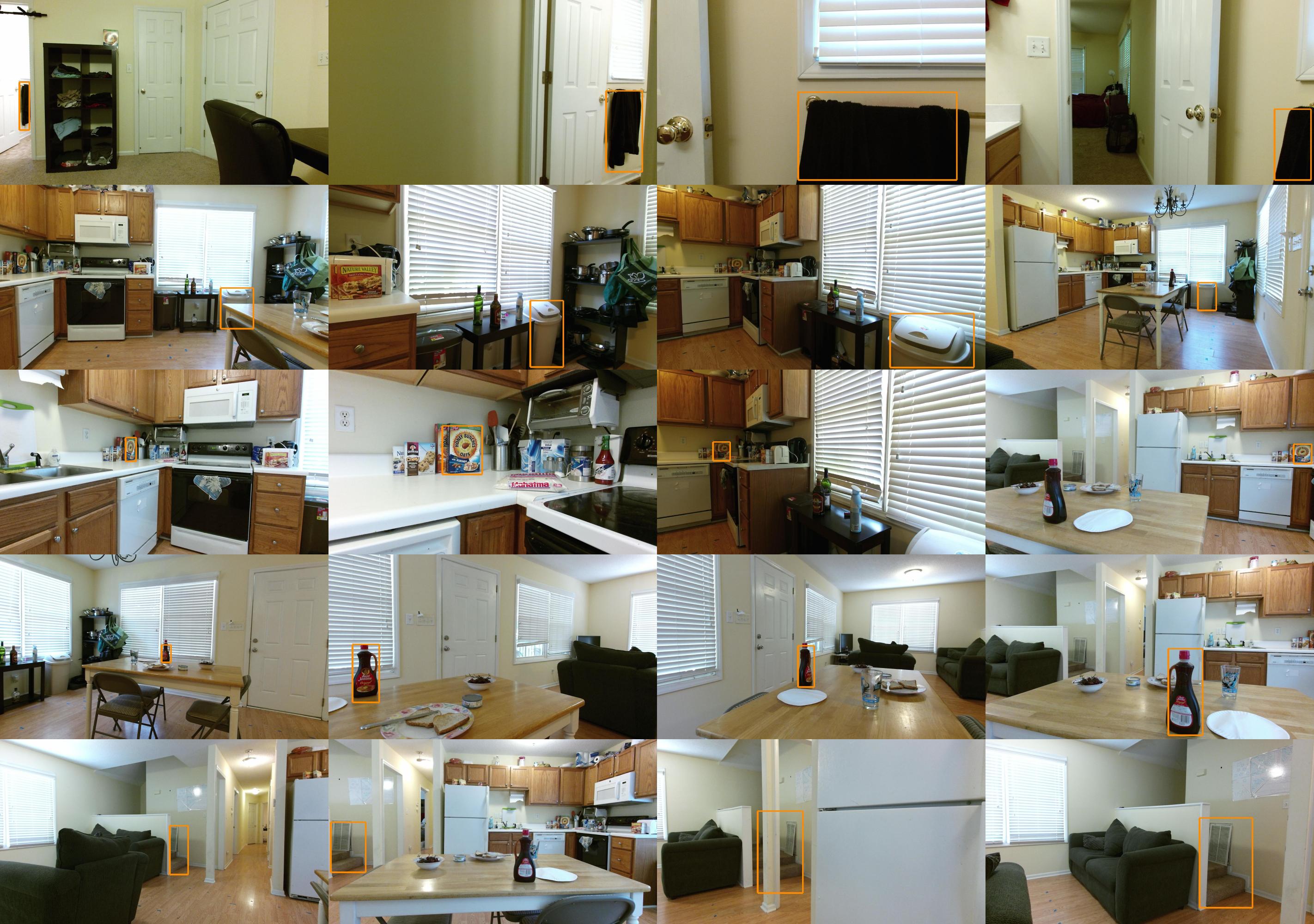}
    \caption{Example of clustering results. Each row correspond to one cluster. From top to bottom: A towel, a bin, a cereal box, a syrup bottle and the stairs. We see that the clusters capture object instances from different viewpoints, even when the instance is partially occluded.}
    \label{fig:clusters}
\end{figure}

Examples of discovered objects are shown in Fig.~\ref{fig:clusters}. Each cluster contains images of the same object from a diverse set of view points. This method discovers not only objects from COCO or AVD, but also other objects and landmarks present in the environments.

\begin{figure}[h]
    \centering
    \includegraphics[width=0.45\textwidth]{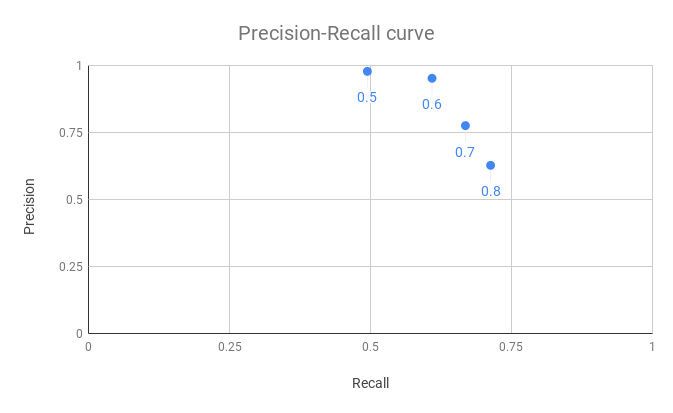}
    \includegraphics[width=0.45\textwidth]{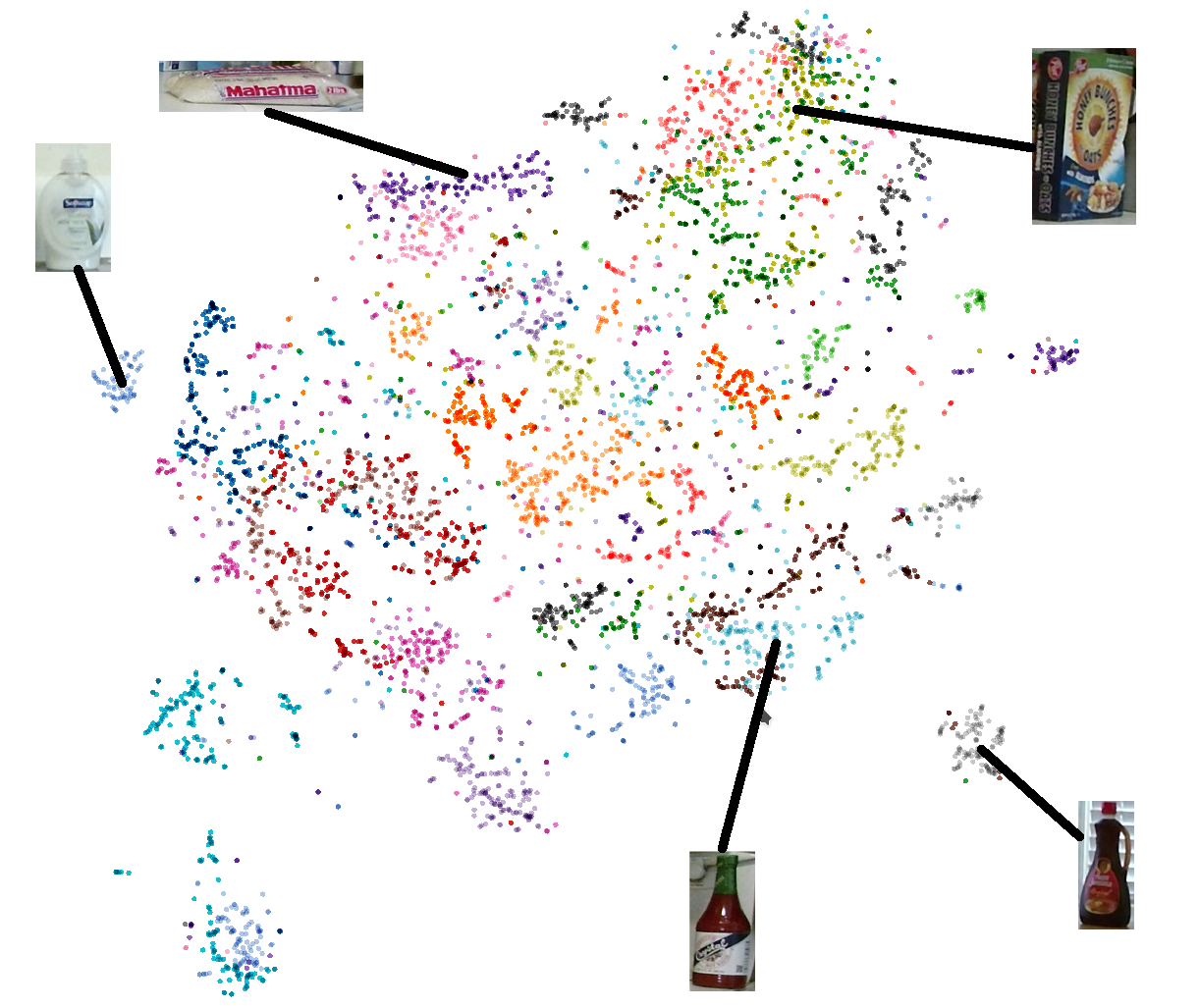}
    \caption{Left: Precision-recall curve for different mean-shift bandwidth values, show below each dot. Right: A 2D T-SNE visualization of the embedding space computed on the ground truth bounding boxes from the testing worlds. Each color correspond to one labeled instance.}
    \label{fig:mean_shift_score}
    \label{fig:tsne}
\end{figure}

Labeled objects may be split within several clusters. To quantify the performance of the object discovery, we measure how well the obtained clusters match the labeled objects from AVD. More precisely, for each of the 33 object instances, we find the dominant cluster which contains the most of this object and compute precision and recall for it. Precision is defined as the portion of the bounding boxes in the cluster covering the object. Recall is the portion of object bounding boxes in our data contained in the cluster. Average precision / recall for different bandwidths is shown in Figure \ref{fig:mean_shift_score}, left.

By considering the biggest clusters for every object instance, we see that each cluster mainly covers a single object ($> 90\%$ of the object examples) and contains majority of the total proposals of this object ($> 60\%$ recall).

In addition to clusters corresponding to some of the original 80 COCO classes, the proposed approach discovers instances of non-COCO classes (see Fig.~\ref{fig:clusters}). The clusters capture objects from a diverse set of viewpoints and is robust to partial occlusion of the object.
Discovery can also be visualized using a T-SNE embedding (Fig.~\ref{fig:tsne}, right). Self-supervisory signals are effective for learning an embedding space in which the representations of the same object instance are successfully clustered together.

\subsection{Object Detection}  \label{sec:experiment_few_shot}
The second application of the self-supervision is object detection. We evaluate the two approaches described in Section~\ref{few_shot}. All methods use Faster-RCNN~\cite{ren2015faster} with the same ResNet-101~\cite{he2016deep} backbone, pre-trained on COCO~\cite{lin2014microsoft}. The feature-extraction part of the network is kept frozen.

\vspace{-0.3cm}
\paragraph{SSOD-Cluster} 
We manually assign labels to the discovered clusters and use them to fine-tune the Region Proposal Network (RPN) as well as the box classification and localization layers of the second stage of Faster-RCNN. In our experiment only the dominant cluster for each instance is used as training data.

\vspace{-0.3cm}
\paragraph{SSOD-Dist} 

We randomly select 1, 3, 5, and 10 bounding box examples per object from the training environment, and classifying every object proposal from the test environment using nearest neighbour classifier. We aggregate the results by averaging the mAP per instance over training on all train / test environments. When computing the mAP, the detection score we use is $2.0 - mindist$, with $mindist$ being distance to the closest labeled embedding, as defined in Section \ref{few_shot}. Thus the bounding boxes above a certain distance from every labeled embedding will be considered as background.

We compare our SSOD variants with a regular Faster-RCNN fine-tuned on the same examples as SSOD-Dist. Faster-RCNN was trained using the Tensorflow object detection API \cite{Huang_CVPR17}, which regularizes the model by performing data augmentations, allowing training with few examples. The model converges after 5000 iterations with batch size of 1. Although the three methods use the same detection model, pre-trained on the same data, fine-tuning differs, as summarized in Table \ref{tab:training_setup}. Since the proposals are fixed for SSOD-Dist, the proposal mechanism cannot be adapted while both the baseline and SSOD-Cluster can fine-tune the first stage of the detector.

The three methods differ in the type and scale of the needed human labeling (second row in Table \ref{tab:training_setup}). For the baseline and SSOD-Dist we require bounding box labels. Since labeling is a bottleneck in robotic scenarios, we can label only a limited number of examples, which places us in a few shot regime. For SSOD-Cluster, however, the boxes used are generated by the RPN and only object labels are assigned per cluster. Thus, a larger number of examples can be labeled, without the need to manually draw bounding boxes.

\begin{table}[h]
    \centering
    \begin{tabular}{|c|c|c|c|}
    \hline
    
- & Faster-RCNN & SSOD-Dist & SSOD-Cluster  \\
\hline
RPN Proposals & Fine tuned & Fixed & Fine tuned \\
\hline
Labelling & Box and Label & Box and Label & Label only \\
    
    \hline
    \end{tabular}
    \caption{Difference of training setup between the experiments.}
    \label{tab:training_setup}
\end{table}

All the methods are compared in Table~\ref{tab:map_per_instance}. For SSOD-Cluster, an average of 170 clusters are labeled per training environment. If we are to use the same labeling budget to label examples for the 30 object instance, we can label approximately 5 examples per object. We see that SSOD approaches outperform the baseline in this regime, with SSOD-Cluster having a substantial boost. 

\begin{table}[h]
    \centering
    \begin{tabular}{|c|c|c|c|}
    
\hline
Instances & F.~RCNN &  \multicolumn{2}{|c|}{SSOD} \\
\cline{3-4}
{} & {} & Dist & Cluster  \\
\hline
can & 0.22 & 0.31 & 0.25  \\
soap & 0.42 & 0.40 & 0.55  \\
bottle & 0.44 & 0.43 & 0.50  \\
other & 0.43 & 0.46 & 0.51  \\

\hline
Average & 0.36 & 0.38 & 0.42  \\
\hline

    \end{tabular}
    \hspace{0.8cm}
    \begin{tabular}{|c|c|c|}
    \hline
    
Train & \multirow{2}{*}{F.~RCNN} & \multirow{2}{*}{SSOD-Dist}  \\
size & {} & {}  \\
\hline
1 & 0.120 & 0.226  \\
3 & 0.210 & 0.264  \\
5 & 0.256 & 0.277  \\
10 & 0.294 & 0.281  \\
All & 0.396 & -  \\
    
    \hline
    \end{tabular}
    
    \caption{Mean Average Precision at $IoU=0.5$. \textit{Left:} Faster RCNN vs SSOD, broken down by coarse object categories. We use 5 labeled examples per object for training F.~RCNN and SSOD-Dist. We exclude rice and cereal boxes as all methods, including baseline, perform poorly. \textit{Right:} Fewshot object detection. We show mAP for different number of training examples per class.}
    \label{tab:map_per_instance}
\end{table}

In the few-shot settings (see Table \ref{tab:map_per_instance}, right) our method achieves a better precision than a standard object detector when the number of training examples is reduced, even though the RPN proposals for the distance metric learning are not fine tuned for those specific instances. The improvement is substantial when we can afford 5 or fewer examples per object, which is realistic assumption when deploying a robot to a new environment. The upper limit of the performance is only $0.40$, which is when we label all $5000$ images in the training environment. SSOD-Dist provides performance and efficiency in training with a very limited labelling budget.

\begin{figure}
    \centering
    \includegraphics[width=0.7\textwidth]{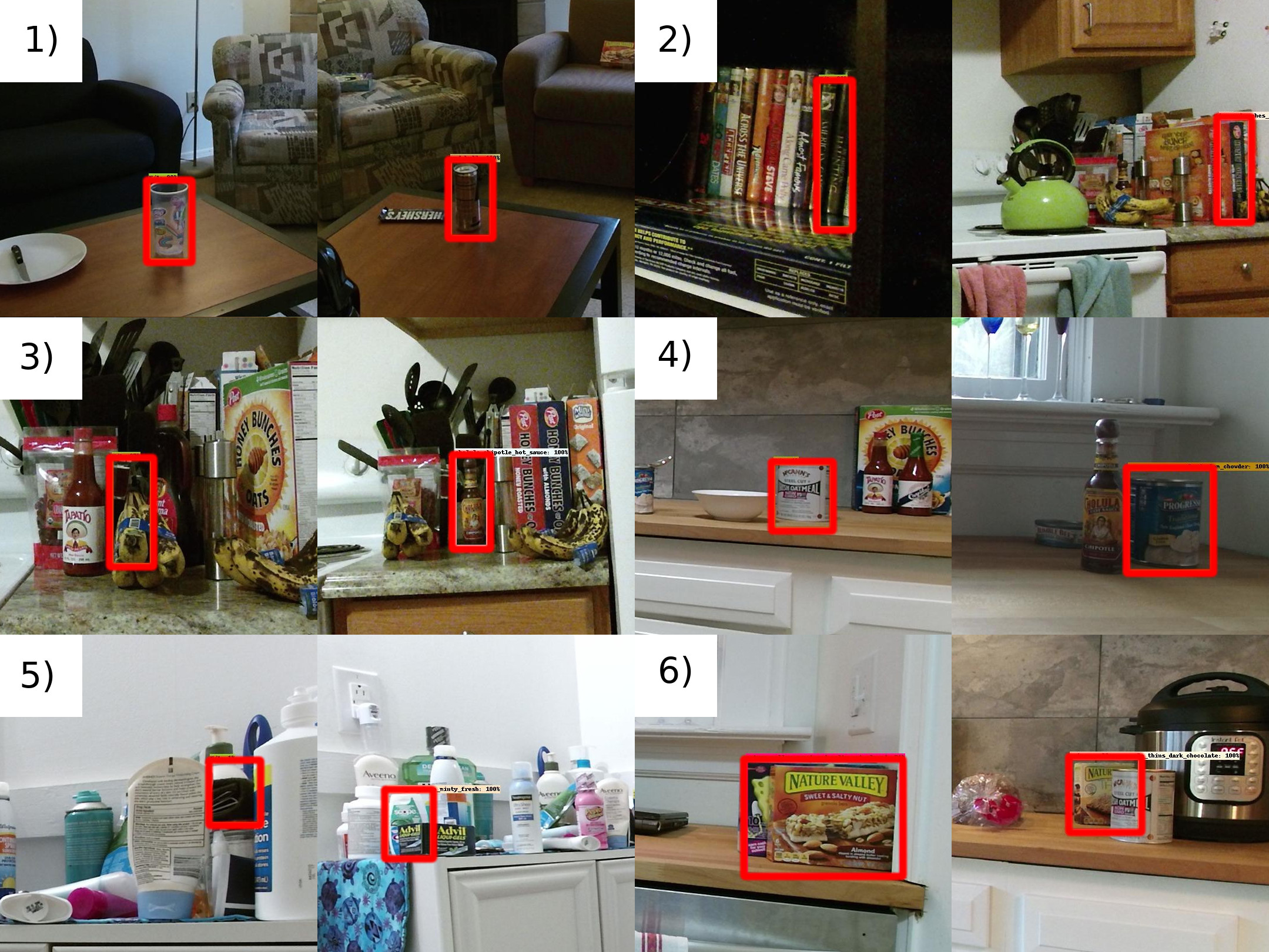}
    \caption{Example of wrong matching between the query (left) and the labeled example (right).}
    \label{fig:dist_failure}
\end{figure}

For all methods, there is disparity in performance between classes. For instance, models performs well for bottle shaped objects, yet have poor performance for cereal boxes. This can be explain as during training, cereal boxes are seen only in a particular orientations and often hidden behind other objects, making it difficult to generalize to new view points. Furthermore cereal boxes have often similar textures and can be confused with each other. Figure \ref{fig:dist_failure} display some examples of incorrect matching for the SSOD-Dist approach, where the bounding box proposal is incorrectly matched with one of the labeled ground truth bounding box. Most mistakes occur because of similar shape/object type (images 1, 2, 4 and 6), texture/colors (images 2, 5 and 6) and location (images 1, 3, 5). Two objects are classified together usually because of a combination of those factor (for instance two objects of similar shape at a similar location as in image 1).

\section{Conclusion}
In this paper, we propose to use self-supervision to learn object detection models. We present several applications: unsupervised object discovery and two efficient object detection models, called SSOD, with strong performance in limited supervision regime.

The proposed approach is especially useful in robotic scenarios, where a robot is to be efficiently deployed to a novel environment. SSOD allows for such an agent to be able to quickly learn new objects and be able to detect them. 

\bibliographystyle{splncs}
\bibliography{nips_2018}

\begin{thebibliography}{10}

\bibitem{schroff2015facenet}
Schroff, F., Kalenichenko, D., Philbin, J.:
\newblock Facenet: A unified embedding for face recognition and clustering.
\newblock In: Proceedings of the IEEE conference on computer vision and pattern
  recognition. (2015)  815--823

\bibitem{ren2015faster}
Ren, S., He, K., Girshick, R., Sun, J.:
\newblock Faster r-cnn: Towards real-time object detection with region proposal
  networks.
\newblock In: Advances in neural information processing systems. (2015)  91--99

\bibitem{sivic2005discovering}
Sivic, J., Russell, B.C., Efros, A.A., Zisserman, A., Freeman, W.T.:
\newblock Discovering objects and their location in images.
\newblock In: Computer Vision, 2005. ICCV 2005. Tenth IEEE International
  Conference on. Volume~1., IEEE (2005)  370--377

\bibitem{russell2006using}
Russell, B.C., Freeman, W.T., Efros, A.A., Sivic, J., Zisserman, A.:
\newblock Using multiple segmentations to discover objects and their extent in
  image collections.
\newblock In: Computer Vision and Pattern Recognition, 2006 IEEE Computer
  Society Conference on. Volume~2., IEEE (2006)  1605--1614

\bibitem{Uijlings_IJCV13}
Uijlings, J., Sande, K., Gevers, T., Smeulders, A.:
\newblock Selective search for object recognition.
\newblock In: in International Journal of Computer Vision (IJCV). (2013)

\bibitem{Collet_ICRA16}
Collet, A., Martinez, M., , Srinivasa, S.:
\newblock The moped framework: Object recognition and pose estimation for
  manipulation.
\newblock in {IEEE} International Conference on Robotics and Automation (ICRA)
  (2016)

\bibitem{Mishra_ICRA12}
Mishra, A., Shrivastava, A., Aloimonos, Y.:
\newblock Segmenting ``simple'' objects using rgb-d.
\newblock in {IEEE} International Conference on Robotics and Automation (ICRA)
  (2012)

\bibitem{Karpathy_ICRA13}
Karpathy, A., Miller, S., , Fei-Fei, L.:
\newblock Object discovery in 3d scenes via shape analysis.
\newblock in {IEEE} International Conference on Robotics and Automation (ICRA)
  (2013)

\bibitem{ssd_eccv2016}
Liu, W., Anguelov, D., Erhan, D., Szegedy, C., Reed, S., Fu, C.Y., Berg, A.C.:
\newblock Ssd: Single shot multibox detector.
\newblock In: European Conference On Computer Vision. (2016)

\bibitem{kwak2015unsupervised}
Kwak, S., Cho, M., Laptev, I., Ponce, J., Schmid, C.:
\newblock Unsupervised object discovery and tracking in video collections.
\newblock In: Computer Vision (ICCV), 2015 IEEE International Conference on,
  IEEE (2015)  3173--3181

\bibitem{pathak2017learning}
Pathak, D., Girshick, R., Doll{\'a}r, P., Darrell, T., Hariharan, B.:
\newblock Learning features by watching objects move.
\newblock In: Proc. CVPR. Volume~2. (2017)

\bibitem{schulter2013unsupervised}
Schulter, S., Leistner, C., Roth, P.M., Bischof, H.:
\newblock Unsupervised object discovery and segmentation in videos.
\newblock In: BMVC, Citeseer (2013)  1--12

\bibitem{agrawal2015learning}
Agrawal, P., Carreira, J., Malik, J.:
\newblock Learning to see by moving.
\newblock In: Computer Vision (ICCV), 2015 IEEE International Conference on,
  IEEE (2015)  37--45

\bibitem{wang2015unsupervised}
Wang, X., Gupta, A.:
\newblock Unsupervised learning of visual representations using videos.
\newblock arXiv preprint arXiv:1505.00687 (2015)

\bibitem{wang2014video}
Wang, L., Hua, G., Sukthankar, R., Xue, J., Zheng, N.:
\newblock Video object discovery and co-segmentation with extremely weak
  supervision.
\newblock In: European Conference on Computer Vision, Springer (2014)  640--655

\bibitem{pathak2017curiosity}
Pathak, D., Agrawal, P., Efros, A.A., Darrell, T.:
\newblock Curiosity-driven exploration by self-supervised prediction.
\newblock In: International Conference on Machine Learning (ICML). Volume 2017.
  (2017)

\bibitem{pinto2016supersizing}
Pinto, L., Gupta, A.:
\newblock Supersizing self-supervision: Learning to grasp from 50k tries and
  700 robot hours.
\newblock In: Robotics and Automation (ICRA), 2016 IEEE International
  Conference on, IEEE (2016)  3406--3413

\bibitem{mitash2017self}
Mitash, C., Bekris, K.E., Boularias, A.:
\newblock A self-supervised learning system for object detection using physics
  simulation and multi-view pose estimation.
\newblock In: Intelligent Robots and Systems (IROS), 2017 IEEE/RSJ
  International Conference on, IEEE (2017)  545--551

\bibitem{hickson2018object}
Hickson, S., Angelova, A., Essa, I., Sukthankar, R.:
\newblock Object category learning and retrieval with weak supervision.
\newblock arXiv preprint arXiv:1801.08985 (2018)

\bibitem{sermanet2017time}
Sermanet, P., Lynch, C., Hsu, J., Levine, S.:
\newblock Time-contrastive networks: Self-supervised learning from multi-view
  observation.
\newblock arXiv preprint arXiv:1704.06888 (2017)

\bibitem{Held_ICRA16}
Held, D., Savarese, S., Thrun, S.:
\newblock Deep learning for single-view instance recognition.
\newblock in {IEEE} International Conference on Robotics and Automation (ICRA)
  (2016)

\bibitem{lin2014microsoft}
Lin, T.Y., Maire, M., Belongie, S., Hays, J., Perona, P., Ramanan, D.,
  Doll{\'a}r, P., Zitnick, C.L.:
\newblock Microsoft coco: Common objects in context.
\newblock In: European conference on computer vision, Springer (2014)  740--755

\bibitem{durrant2006simultaneous}
Durrant-Whyte, H., Bailey, T.:
\newblock Simultaneous localization and mapping: part i.
\newblock IEEE robotics \& automation magazine \textbf{13}(2) (2006)  99--110

\bibitem{comaniciu2002mean}
Comaniciu, D., Meer, P.:
\newblock Mean shift: A robust approach toward feature space analysis.
\newblock IEEE Transactions on pattern analysis and machine intelligence
  \textbf{24}(5) (2002)  603--619

\bibitem{ammirato2017dataset}
Ammirato, P., Poirson, P., Park, E., Ko{\v{s}}eck{\'a}, J., Berg, A.C.:
\newblock A dataset for developing and benchmarking active vision.
\newblock In: Robotics and Automation (ICRA), 2017 IEEE International
  Conference on, IEEE (2017)  1378--1385

\bibitem{bigbird2014}
Singh, A., Sha, J., Narayan, K., Achim, T., Abbeel, P.:
\newblock A large scale 3d database of object instances.
\newblock In: IEEE Conference on Robotics \& Automation, IEEE (2018)

\bibitem{he2016deep}
He, K., Zhang, X., Ren, S., Sun, J.:
\newblock Deep residual learning for image recognition.
\newblock In: Proceedings of the IEEE conference on computer vision and pattern
  recognition. (2016)  770--778

\bibitem{Huang_CVPR17}
Huang, J., Rathod, V., Sun, C., Zhu, M., Korattikara, A., Fathi, A., Fischer,
  I., Wojna, Z., Song, Y., Guadarrama, S., Murphy, K.:
\newblock Speed/accuracy trade-offs for modern convolutional object detectors.
\newblock Computer Vision and Pattern Recognition (CVPR) (2017)

\end{thebibliography}

\end{document}